# An algorithm for Left Atrial Thrombi detection using Transesophageal Echocardiography

Jianrui Ding[a], Min Xian[b], H.D. Cheng[b,*], Yang Li[c], Fei Xu[b], Yingtao Zhang[a]

[a]School of Computer Science and Technology, Harbin Institute of Technology, Harbin, 150001, PR China

[b]Department of Computer Science, Utah State University, Logan, UT 84322, USA

[c]Fourth Affiliated Hospital of Harbin Medical University, Harbin, 150001, PR China

Abstract：

Transesophageal echocardiography (TEE) is widely used to detect left atrium (LA)/left atrial appendage (LAA) thrombi. In this paper, the local binary pattern variance (LBPV) features are extracted from region of interest (ROI). And the dynamic features are formed by using the information of its neighbor frames in the sequence. The sequence is viewed as a bag, and the images in the sequence are considered as the instances. Multiple-instance learning (MIL) method is employed to solve the LAA thrombi detection. The experimental results show that the proposed method can achieve better performance than that by using other methods.

# 1. Introduction

About 20%–40% of stroke cases are caused by cardiogenic emboli [1]. The left atrial appendage (LAA) is an important location to form thrombus which can cause subsequent cardio embolic events, such as atrial fibrillation (AF) [2] which is a relatively common supraventricular arrhythmia. It tends to develop a thrombus in the left atrium (LA)/left atrial appendage (LAA) when it lasts for more than 48 hours [3].

Transesophageal echocardiography (TEE) is considered as a sensitive tool to detect LAA thrombi [4]. Although TEE is widely used, it is semi-invasive and requires special skills for proper performance. In addition, due to the complex anatomy and substantial variations of the LAA, there are some artifacts in the image. Its interpretation requires well-trained radiologists, and even experts may have inter-observer variation. Operation dependency and poor image quality may lead to misdiagnoses of LAA thrombi.

Computer-aided diagnosis (CAD) has been widely used in medicine, such as breast tumor [5] and thyroid nodules [6] detection. CAD can provide quantitative and objective information for clinicians to reduce diagnosis variability. It has the ability to excavate some characteristics in image which are indistinct for radiologists. And it can give "a second opinion" to support clinical decision-making. It has been proved that CAD can help radiologists to detect tumor lesions effectively and accurately. However, to our knowledge, there is little research on CAD for detecting LA/LAA thrombi by TEE [8].

Texture analysis has already demonstrated considerable potential in medical image analysis for lesion segmentation and characterization [5, 9]. Dynamic texture is a spatially repetitive, time-varying visual pattern in an image sequence [10], which can be utilized for medical image sequence analysis [11]. Local binary patterns (LBPs) are widely used descriptors for texture analysis [14] which have been extensively studied in medical field [15, 16].

On TEE image, a thrombus appears as a well-circumscribed, uniformly consistent echogenic mass whose texture is different from that of the LAA wall [7], as shown in Fig.1. TEE is a dynamic process; and the image sequence is recorded for analysis by radiologists. The texture can be used to characterize the thrombus for differentiating it from other tissues, while the dynamic texture can be used to characterize the variation of thrombus in the sequence.

Multiple-instance learning (MIL) was proposed to solve learning problems with incomplete information about the data labels [12] and is widely employed for many applications including medical image analysis [5]. Radiologist usually labels a TEE sequence as thrombus or non-thrombus, but, the individual images in the sequence have no labels. Hence, the problem of TEE image classification can be viewed as a MIL problem.

In the previous work [8], five images were selected from each sequence. The images were enhanced in frequency domain by applying Gaussian low-pass filter to remove noise. The co-occurrence matrix was utilized to describe the texture information of the region of interest (ROI) marked by radiologist. Then an artificial neural network (ANN) was used to classify the ROI as thrombi or muscle.

For extracting texture features of ROI, three distances (d = 1, 2, 3), and four directions (= 0°, 45°, 90°, 135°) were used to compute the co-occurrence matrices. From each co-occurrence matrix, four descriptors were calculated: contrast, correlation, energy, and homogeneity. In addition, entropy, mean and standard deviation of ROI were calculated.

The co-occurrence features were also calculated from the ROIs of its neighboring images in the same sequence. The dynamic texture of a ROI was formed by these three co-occurrence feature sets.

Finally, an ANN was trained by the data. For a new sequence, the five ROIs were classified by the ANN. And the results were voted, i.e., if three or more ROIs were classified as thrombi, the sequence was recognized as thrombus.

In this paper, we will improve the method by further considering dynamic texture features and sequence classification problem; and propose a novel method for sample selection to remove outliers.

The appearance of lesion varies among images in the sequence, such as location, shape, margin, texture, etc. Such variations are useful for radiologist to catch the salient feature of the lesion. Dynamic texture feature is employed in the paper to model the variations and to reflect the dynamic change of the lesion in the sequence effectively.

An image sequence is labeled as thrombus or non-thrombus, but the label of each image in the sequence is unknown. If we consider all images of the sequence labeled as thrombus are thrombus, and all images of the sequence labeled as muscle are muscles; all these images participate in training process. But such assumption may introduce unlabeled samples into learning phase; these samples are considered as noise or outliers to affect the accuracy of the model. We use a novel approach to select the samples from the sequence to avoid outlier problem.

The description of our method is as follows. The image is transformed to frequency domain and Gaussian low-pass filter is applied to remove noise. Then non-linear transform is utilized to enhance the contrast. After image enhancement the ROI is characterized by local binary pattern variance (LBPV) histogram [17] features considering both global information and local texture features. Then the dynamically varying characteristic is utilized to model the dynamic texture feature of a ROI. A sequence can be viewed as a bag and the images in the sequence as the instances. MIL assumption and a novel approach for removing outliers is employed to build the training samples. Finally, the support vector machine (SVM) is trained by the samples to classify the sequences. The flowchart of the proposed approachis presented in Fig. 2. The experimental results confirm that the performance of the proposed method is much better than that of the existing methods.

**2. Materials and Methods**

2.1 Image acquisition

The TEE sequences in this study contain 264 cases (114 thrombi and 150 muscles) which were collected by using iE33 xMatrix echocardiographic system (Philips Healthcare, Bothell, WA) equipped with a 2–7-MHz matrix array in the TEE probe (X7-2t; Philips Healthcare) by the First Affiliated Hospital of Harbin Medical University. All the diagnoses were made according to the standard criteria [18] and further confirmed surgically or clinically. The study protocol was approved by the Institutional Review Board of the First Affiliated Hospital of Harbin Medical University, and written informed consent was obtained from each patient. The privacy of the patients has been well protected.

Five images are selected from each sequence by the radiologist with the following criteria: (1) the LAA was clearly shown in the image; and (2) the LAA was in diastole so that thrombi could be detected [8]. And the ROI of every image is marked by the radiologist. Total 1320 images are acquired.

2.2 Image enhancement

Image enhancement was performed to remove noise and to enhance contrast. First, the image is transferred from space domain to frequency domain by using Fast Fourier Transform. Then, Gaussian low-pass filter was applied to remove the noise and smooth the image. Finally, a nonlinear enhancement operation was used by keep the gray values of the area with high echoes and attenuate the low echoes to improve the contrast of the smoothed image [8]. The flowchart of the image enhancement method is presented in Fig. 3.

Gaussian low-pass filter is defined as follows:

$$G(u,v) = e^{-D^2(u,v)/2\sigma^2} \tag{1}$$

where $D(u,v) = \sqrt{u^2 + v^2}$, is the distance from the origin in the frequency plane, $\sigma$ measures the spread or dispersion of Gaussian curve.

The nonlinear enhancement operation is defined:

$$T(x,y) = We^{-[(g(x,y)-\max(g(x,y))]^2/B} \quad (2)$$

where $W$ and $B$ are constants.

An example is presented in Fig. 4.

2.3 Texture feature extraction

Texture is an important feature for image analysis, and has been used in many applications, such as medical imaging, material identification and image classification [18]. One of the commonly used texture descriptors is the local binary pattern (LBP) operator due to the following facts: its resistance to lighting changes, low computational complexity, and ability to code the fine details [14]. Local binary pattern variance (LBPV) is a variant of LBP which integrates local contrast information into LBP. The LBPV histogram features are extracted to describe the ROI.

The LBPV histogram is defined as follows [17]:

$$LBPV_{P,R}^{riu2}(k) = \sum_{i=1}^{N}\sum_{j=1}^{M} mask(i,j) \cdot w(LBP_{P,R}^{riu2}(i,j), k), \quad k \in [0,K] \quad (3)$$

$$w(LBP_{P,R}^{riu2}(i,j), k) = \begin{cases} VAR_{P,R}(i,j), & LBP_{P,R}^{riu2}(i,j) = k \\ 0 & \text{otherwise} \end{cases} \quad (4)$$

$$mask(i,j) = \begin{cases} 1, & (i,j) \in ROI \\ 0, & \text{otherrwise} \end{cases} \quad (5)$$

$$LBP_{P,R}^{riu2} = \begin{cases} \sum_{p=0}^{P-1} s(g_p - g_c) & \text{if } U(LBP_{P,R}) \leq 2 \\ P+1 & \text{otherwise} \end{cases} \quad (6)$$

$$s(x) = \begin{cases} 1, & x \geq 0 \\ 0, & x < 0 \end{cases} \tag{7}$$

$$VAR_{P,R} = \frac{1}{P}\sum_{p=0}^{P-1}(g_p - \mu)^2 \tag{8}$$

$$\mu = \frac{1}{P}\sum_{p=0}^{P-1} g_p \tag{9}$$

where $P$ is the number of neighbors, $R$ is the radius of the neighborhood, $K$ is number of the bins, $g_p$ is the gray value of its neighbor, and $g_c$ is the gray value of the center pixel. The value of $U$ is the number of spatial transitions (bitwise 0/1 changes).

2.4 Dynamic texture feature construction

Dynamic textures exhibit some forms of temporal stationary, such as waves, steam, and foliage [19] of the images in the sequences. In our study, image sequences are acquired and the variation of features among images can be applied to describe the lesions.

The *LBPV* features are extracted from the ROI, denoted as $F_R$. The *LBPV* features are also extracted from the neighboring images in the same sequence, denoted as $F_{R-}$ and $F_{R+}$, respectively. Comparing with muscle, the texture of thrombus varies less. Then the difference of the features is used to form the dynamic texture as following:

$$F = [F_R ; \frac{(abs(F_{R+} - F_R) + abs(F_R - F_{R-}))}{2}] \tag{10}$$

The new feature vector $F$ is composed of two parts, one is the feature of ROI denoted as $F_R$, and the other is the difference feature of its neighbors denoted as $\frac{(abs(F_{R+} - F_R) + abs(F_R - F_{R-}))}{2}$. Then, $F$ can reflect the feature of ROI and the variation of the ROI as well. The features are normalized to [0, 1] before input to a classifier.

An example is presented in Fig. 5.

The features of the three images in a sequence are combined to produce the dynamic features.

2.5 Bag and instances construction

Multiple-instance learning (MIL) is proposed to solve learning problems with incomplete information about the labels of the data. For traditional supervised learning, each training example is represented by a fixed-length feature vector with known label. However, in MIL each example is called a bag and represented by multiple instances. The number of instances in each bag can be different. In other words, bags are represented by variable-length vectors. Labels are only provided for the training bags, and the labels of instances are unknown. The MIL task is to learn a model to classify new bags [20]. The key challenge of MIL is the ambiguity in the labels of the instances.

Here, for our problem, each sequence has five images. The sequence is labeled as the thrombus or muscle; however, the labels of the images in the sequence are unknown. The sequence can be considered as a bag and the images can be viewed as the instances; then the problem is transformed into a MIL problem.

We adapt the assumption of the standard MIL: the positive bag has at least one positive instance, and the negative bag has no positive instances. That is to say, for a sequence which is labeled as thrombus, there is at least one image classified as thrombus; while for a sequence which is labeled as muscle, all of the images in the sequence are classified as muscle.

To decide which instance is thrombus and which instance is muscle, we define the feature center of the muscle as below:

$$M_c = \frac{1}{N}\sum_{i=1}^{N} F_i \tag{11}$$

where $N$ is the total number of instances of muscle, and $F_i$ is the texture feature of an instance belonging to a sequence labeled as muscle.

And the representing instance in a sequence labeled as thrombus is defined as the furthest instance from the muscle center obtained by Eq. (9).

$$F_{t_i} = max\left(dist\left(F_{t_j}, M_c\right)\right) \tag{12}$$

where $i$ is the $i$th sequence whose label is thrombus, and $j$ is the $j$th ROI in the sequence.

Therefore, an instance which is considered as the thrombus is selected from the sequence labeled as the thrombus. While all the instances in the sequence labeled as muscle are viewed as muscle. To avoid the unbalance problem during the training and classification, the muscle sequence is represented by the average of the muscle instances in the sequence as Eq. (11).

$$F_{m_i} = \frac{1}{M}\sum_{j=1}^{M} F_{m_j} \tag{13}$$

where $i$ is the $i$th sequence whose label is muscle and $j$ is the $j$th ROI in the sequence, $M$ is the number of ROIs selected from the sequence. For our problem, $M = 5$.

2.6 Training and classification

After instance selection, a traditional supervised classifier, SVM, can be trained and utilized to classify the sequences into thrombus or muscle.

Support vector machine (SVM) is a widely used pattern classification technique. SVM aims at maximizing the margin between the separating hyper plane and the data to minimize the upper bound of generalization error. Unlike traditional methods which minimize the empirical training error, SVM can be regarded as an approximate implementation of the Structure Risk Minimization principle [21]. Property of condensing information in the training data and providing a sparse representation by using a very small number of data points (Support Vectors) makes SVM very attractive.

The training samples are mapped to higher dimension space with a kernel function, and an optimal decision plane can be created [22]. The binary classification problem can be solved by minimizing the function below by SVM.

$$\Phi(w, \xi) = \frac{1}{2}(w * w) + C(\sum_{l=1}^{L} \xi_l) \qquad (14)$$

where:

$$\forall \xi_l \geq 0, w * x_l + b \geq 1 - \xi_l \: if \: y_l = 1$$

and

$$w * x_l + b \leq -1 + \xi_l \: if \: y_l = -1$$

where $w$ is the separating plane to be solved, $\xi$ is the soft margin, $x_l$ is the training sample, $y_l$ is the known label of $x_l$, $L$ is the number of training samples, and $C$ is a constant.

The problem can be transformed to find the parameter vector $\alpha^0$ to maximize the following function using Lagrange method.

$$w(\alpha) = \sum_{l=1}^{L} \alpha_l - \frac{1}{2} \sum_{i,j}^{L} \alpha_i \alpha_j y_i y_j K(x_i, x_j)$$

$$subject\ to\ \sum_{l=1}^{L} \alpha_l y_l = 0,\ 0 \leq \alpha_l \leq C \tag{15}$$

where $K(\cdot,\cdot)$ is the kernel function. For each training sample $x_i$, there is a corresponding parameter $\alpha_i^0$, if $\alpha_i^0 \neq 0$ the training sample is called a support vector.

After training, for a test sample $x$, its class can be determined by the function below:

$$F(x, \alpha^0) = sign\left(\sum_{l=1}^{N_{SV}} y_l \alpha_l^0 K(x, x_S) + b\right) \tag{16}$$

where $x_S$ is a support vector, $N_{SV}$ is the number of the support vectors and $K(x, x_S)$ is the kernel function [23].

There are four basic kernels [24]:

linear: $$K(x_i, x_j) = x_i^T x_j \tag{17}$$

polynomial: $$K(x_i, x_j) = (\gamma x_i^T x_j + r)^d,\ \gamma > 0 \tag{18}$$

radial basis function (RBF):

$$K(x_i, x_j) = exp\left(-\gamma \|x_i - x_j\|^2\right),\ \gamma > 0 \tag{19}$$

sigmoid $$K(x_i, x_j) = \tanh(\gamma x_i^T x_j + r) \tag{20}$$

RBF kernel nonlinearly maps samples into a higher dimensional space, unlike the linear kernel; it can handle the case when the relation between class labels and attributes is nonlinear. The polynomial kernel has more hyper parameters than the RBF kernel; hence, has high complexity than RBF kernel. In this paper, RBF kernel is selected as the kernel function.

There are two advantages of SVM: the generalization ability is optimal by maximizing the margin distance, and it can solve nonlinear classification tasks by mapping samples to higher dimension space [23].

## 3. Results

There are five images selected from each sequence by the radiologist. And the ROI of every image is marked by the radiologist. Then total 1320 images are acquired. For image enhancement, $\sigma = 15$, $W = 1$ and $B = 50$ were determined by experiment. When *LBPV* features are extracted, $P = 16$ and $R = 2$ are utilized. Besides the *LBPV* features; the local entropy, mean and standard variance are also extracted from a ROI.

After bags and instances obtained, k-fold cross validation is used. K-fold cross validation is a way to improve over the holdout method. The data set is divided into k subsets, and the holdout method is repeated k times. Each time, one of the k subsets is used as the test set and the other k-1 subsets are together to form the training set. Then the average error across all k trials is computed. K-fold cross validation is a very popular method for classifier performance evaluation, especially, for relatively limited size databases which has validated statistically; therefore; it is commonly employed in machine leaning and medical fields [25-28]. In our experiment, k = 10. All the images are randomly divided into 10 groups. Each time, one group is for testing and the rest groups are for training.

The performance of the proposed feature extraction and classification strategy is evaluated by the classification accuracy. Define the number of correctly and incorrectly classified thrombi as the true positive (TP) and false negative (FN); and the number of correctly and incorrectly classified muscles as true negative (TN) and false positive (FP), respectively. The sensitivity (SE) is defined as: TP / (TP+FN); the specificity (SP) is defined as: TN / (TN+FP); and the classification accuracy (ACC) is defined as: (TP+TN) / (TP+TN+FP+FN).

For determining the parameters C and $\gamma$ of the SVM, the grid search method is used [24]. The aim is to find a good combination of C and $\gamma$, which makes the classifier predict unknown samples accurately. C is choose from the [$2^{-5}$, $2^{-3}$, $2^{-1}$, $2^1$, $2^3$, $2^5$, $2^7$, $2^9$, $2^{11}$, $2^{13}$], and $\gamma$ is choose from [$2^{-15}$, $2^{-13}$, $2^{-11}$, $2^{-9}$, $2^{-7}$, $2^{-5}$, $2^{-3}$, $2^{-1}$, $2^1$, $2^3$]. Various sets of C and $\gamma$ are tried, and the set with the best cross-validation accuracy is selected.

For comparing with the previous method [8], feature extraction and classification algorithm proposed in this paper are validated by experiments. For testing the effectivity of *LBPV* features, *LBPV* features of ROI are extracted, and the *LBPV* features of its neighbor image are also extracted. Then, the features are concatenated to represent the ROI. The SVM classifier was applied using the above features to get classification results. When classifying a sequence, the five images in the sequence are classified and then the voting method is applied. That is, if three or more images are classified as thrombi, then the sequence is classified as thrombus.

For checking the validity dynamic texture features, *LBPV* features of ROI are extracted and dynamic texture features are constructed by utilizing its neighboring image information. Then, the voting method is used for sequence classification.

For verifying rationality of the MIL, after dynamic texture feature construction,

the MIL method is employed for sequence classification.

The classification performances of different methods were evaluated by Receiver Operating Characteristic (ROC) curves. Area under ROC curve (AUC) and the standard errors were derived using the bootstrap resampling approach with 5000 iterations. The Z-test was employed to compare the classification performances as well. The Statistical analysis and graphs were made by SigmaPlot for Windows Version 12.5 (Systat Software, Inc.) A two-tailed $p < 0.05$ was considered to be statistically significant. The results are shown in Table 1 and the ROC curve is presented in Fig. 6.

From Table 1, we can see that the LBPV feature proposed by this paper has higher accuracy than that of [8] (90.15 vs. 87.12, 92.42 vs. 87.12, and 95.46 vs. 87.12). ***And the best performance is reached when MIL integrated with the dynamic features (95.46).*** This confirms that the *LBPV* features can be used to detect thrombi; and the dynamic features can reveal the varying process in the sequence further. For the sequence classification problem, it is more suitable for applying MIL approach than the method in [8].

The area under curve (AUC) of the proposed method is higher than that of the method in [8] (0.9854 vs. 0.9381, $p < 0.001$).

The software runs on a 4-core personal computer (Intel Core i5 CPU, 3.20GHz) with 4 GB of RAM, using MATLAB R2014a. After the classifier is trained (it only needs to be trained once), the test images are processed by enhancing, feature extraction and classification; which takes about 5.32s for one image.

4. **Discussions**

A novel MIL method is proposed to detect thrombi. The results show that our method can help radiologist to improve the detection accuracy of the lesion using TEE by quantitatively and objectively describing the features of thrombi.

Besides the texture feature in still images, the variation during dynamic process in an image sequence is also an important indicator of the lesions. By extracting images from a sequence, the difference between images can be used to characterize the dynamic variation. The training set can be formed by selecting typical images from a sequence. But for the sequence classification problem, the class label usually is assigned to the sequence, not to the individual image in the sequence. Therefore, the labels of selected images are ambiguous. And it can introduce noise to the training set; thus, may affect the classification accuracy. .

The proposed method introduces dynamic LBPV features to characterize the variation of the sequence. And it employed MIL to remove noise in the training set. The advantages of the proposed method are: (1) due to gray level variation and movement, the *LBPV* features are more suitable for describing thrombi; (2) the dynamic feature can describe the dynamic nature of the sequence effectively; (3) the proposed approach in the paper is more suitable for classifying the sequences, and can reduce the noise (outliers) in the training samples. The experimental results demonstrate that the proposed approach can achieve better performance.

Precisely locating the lesion can improve the feature extraction and classification performance. In this paper, the lesion is delineated by experienced radiologists.

Dynamic feature construction is also an important factor of the classification problem. In our method, difference between neighbor images was used. Automated segmentation of the lesions and more effective method for constructing the dynamic features which can better describe thrombi will be studied in the future.

5. **Conclusions**

TEE is a widely used medical imaging technique to detectLA/LAA thrombi. But its evaluation criteria are quite subjective. And they are highly dependent on the radiologists' experience.

The method proposed for thrombi detection is based on MIL, dynamic texture features, and novel approach for sample selection. The experimental results demonstrate that dynamic textures can effectively describe the TEE sequences. The MIL method is more suitable for image sequence classification. The method provides an objective and quantitative evaluation for the thrombi detection using TEE.

Acknowledgments

This work is supported, in part, by National Science Foundation of China; the Grant numbers are 61073128 and 61100097.

Fig. 1.(a) Original image (b) Marked by radiologist

Fig. 2.Flowchart of the proposed method

Fig. 3. Flowchart of the image enhancement

Fig. 4. Image Enhancement

(a) Original image (b) Smoothed image (c) Enhanced image

Fig. 5. Dynamic Features Construction

(a) $(R-1)^{th}$ image    (b) $R^{th}$ image    (c) $(R+1)^{th}$ image

Fig. 6. The ROC curves of the proposed method and the one in [8]

Table 1 classification performance comparing with that of [8]

| Method | SE (%) | SP (%) | ACC (%) | AUC | 95% CI |
|---|---|---|---|---|---|
| Method in [8] | 84.21 | 89.33 | 87.12 | 0.9381 | 0.9075 - 0.9688 |
| LBPV | 86.84 | 92.67 | 90.15 | 0.9370 | 0.9054 - 0.9685 |
| LBPV dynamic feature (LBPVD) | 90.35 | 94 | 92.42 | 0.9596 | 0.9358 - 0.9835 |
| LBPV dynamic feature with MIL | 95.61 | 95.33 | 95.46 | 0.9854 | 0.9726 - 0.9981 |

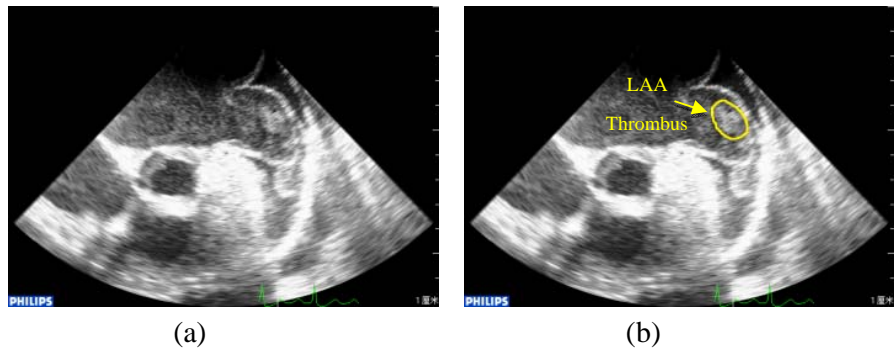

Fig. 1. (a) Original image (b) Marked by radiologist

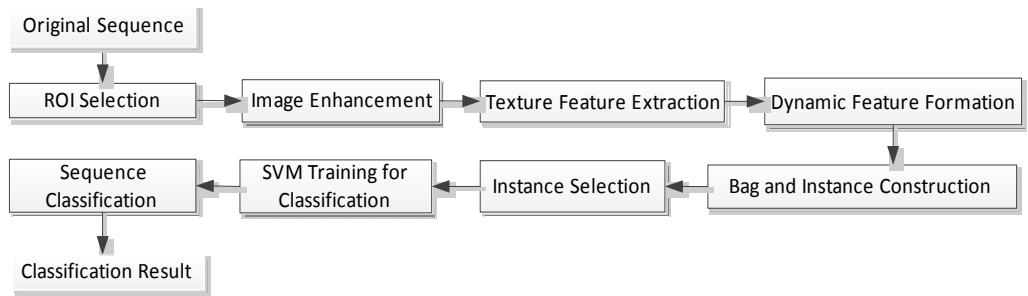

Fig. 2. Flowchart of the proposed method

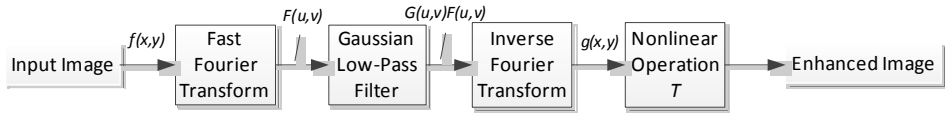

Fig. 3. Flowchart of the image enhancement

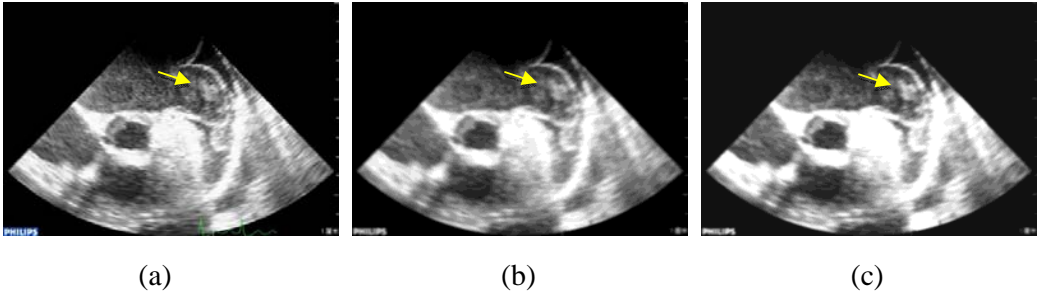

(a)     (b)     (c)

Fig. 4. Image Enhancement

(a) Original image (b) Smoothed image (c) Enhanced image

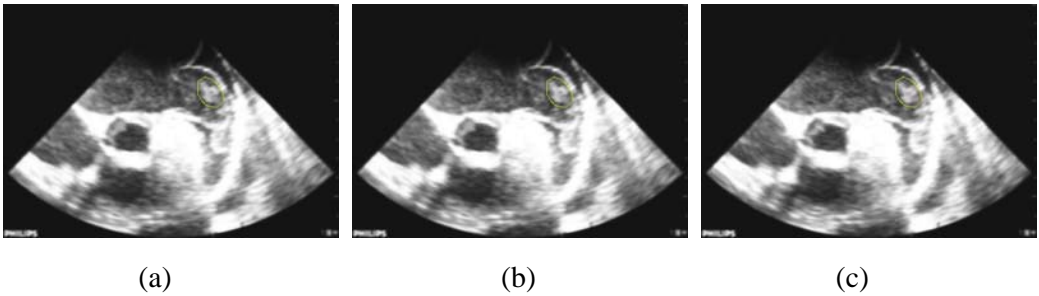

(a)     (b)     (c)

Fig. 5. Dynamic Features Construction

(a) $(R-1)^{th}$ image    (b) $R^{th}$ image    (c) $(R+1)^{th}$ image

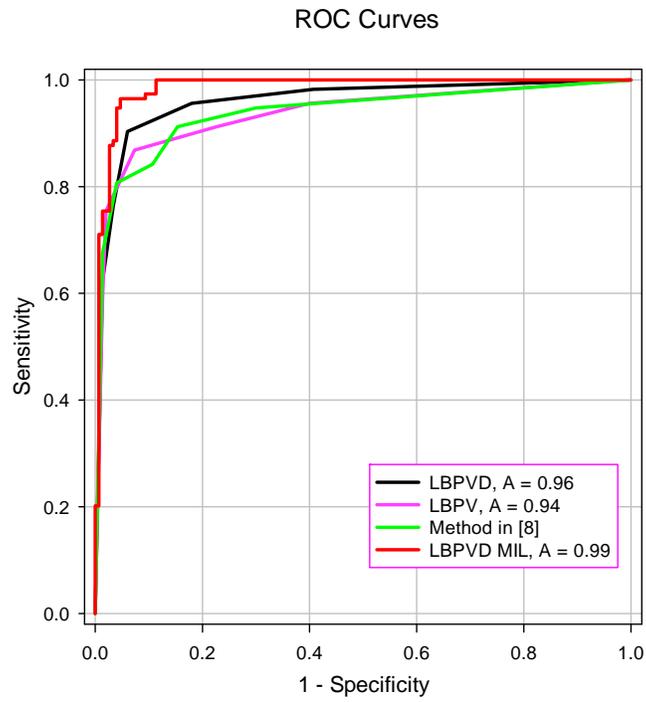

Fig. 6. The ROC curves of the proposed method and the one in [8]